\documentclass[twoside,11pt]{article}

\usepackage{automl2020}

\usepackage{amsmath}
\usepackage{algorithm}
\usepackage[noend]{algpseudocode}
\usepackage{pdfpages}
\usepackage{graphicx}
\graphicspath{ {./images/} }
\usepackage{dblfloatfix}

\jmlrheading{Pedro F da Costa, Romy Lorenz, Ricardo Pio Monti, Emily Jones and Robert Leech}

\ShortHeadings{Bayesian optimization for automatic design of face stimuli}{ da Costa, Lorenz, Pio Monti, Jones, and Leech}
\firstpageno{1}

\begin{document}

\title{Bayesian Optimization for real-time, automatic design of face stimuli in human-centred research} 

\author{\name Pedro F. da Costa \email pedro.ferreira\_da\_costa@kcl.ac.uk \\
       \addr King's College London $\And$ Birkbeck \enspace College,\enspace UK
       \AND
       \name Romy Lorenz \email romy.lorenz@mrc-cbu.cam.ac.uk \\
       \addr University of Cambridge, UK $\And$ Stanford \enspace University,\enspace USA
       \AND
       \name Ricardo Pio Monti \email r.monti@ucl.ac.uk  \\
       \addr Gatsby Computational Neuroscience Unit, UCL, UK
       \AND
       \name Emily Jones \email e.jones@bbk.ac.uk \\
       \addr Birkbeck College, UK
       \AND
       \name Robert Leech \email robert.leech@kcl.ac.uk \\
       \addr King's College London, UK
       }

\maketitle

\begin{abstract}
Investigating the cognitive and neural mechanisms involved with face processing is a fundamental task in modern neuroscience and psychology. To date, the majority of such studies have focused on the use of pre-selected stimuli. The absence of personalized stimuli presents a serious limitation as it fails to account for how each individual face processing system is tuned to cultural embeddings or how it is disrupted in disease. In this work, we propose a novel framework which combines generative adversarial networks (GANs) with Bayesian optimization to identify individual response patterns to many different faces. Formally, we employ Bayesian optimization to efficiently search the latent space of state-of-the-art GAN models, with the aim to automatically generate novel faces, to maximize an individual subject's response. We present results from a web-based proof-of-principle study, 
where participants rated images of themselves generated via performing Bayesian optimization over the latent space of a GAN.
We show how the algorithm can efficiently locate an individual’s optimal face while mapping out their response across different semantic transformations of a face; inter-individual analyses suggest how the approach can provide rich information about individual differences in face processing.
\end{abstract}

\section{Introduction}
Face perception and processing is fundamental for human survival. Within a fraction of seconds, faces reveal to us information about the emotions, gender, age, trustworthiness or intention of another human.
Therefore, faces are among the most important visual stimuli in the natural world and, consequently, a large portion of neuroscience and psychology research has been dedicated to studying its mechanisms \citep{Tsao2008,Eimer2012,Kanwisher1997,Kanwisher2006}. As a result, we now know humans have a specialized neural mechanism to process faces that is tuned by their individual experience \citep{Pascalis2011}. Furthermore, neuroimaging studies have shown that different face stimuli elicit different brain response patterns \citep{Kriegeskorte2007}. This heterogeneity in our neural response to faces presents a challenge to current methodology in the field, where the status quo consists of using the same set of pre-selected face stimuli for every individual and then drawing conclusions from group-level results. Besides not allowing to tailor face stimuli to specific research questions (e.g., what kind of face stimuli maximise response in a given brain region), this approach completely overlooks inter-individual differences in face processing. If we want to better understand the mechanisms underlying face processing, how it develops and how it is disrupted  (e.g., Autism spectrum disorder or fronto-temporal dementia), we need an approach sensitive to individual responses. 

To address this shortcoming, we present a framework that leverages auto-ML and generative neural networks to tailor face stimuli with the aim to maximise an individual subject's response (e.g., neural, behavioural or subjective)\footnote{Code access to the full framework is available at \textit{github.com/PedroFerreiradaCosta/FaceFitOpt}}. By requiring a small number of iterations, this approach bypasses the inherent limitation of participants' attention and familiarity effects from repeated testing. 
Our closed-loop and automated approach measures how the manipulation of face stimuli alters evoked measures. For this, we created a continuous space, where each dimension manipulates a facial semantic attribute orthogonally from the other facial attributes.  We automatically search through this "face space" using Bayesian optimization that queries only the most informative points in that space in order to find the maximum of a target function. The target function can be neural, such as the participant’s brain signal while processing the face stimulus or a behavioural evaluation, such as similarity to a target face or aesthetic judgement. The face stimuli are generated by a generative adversarial network (GAN).

GANs are effective at image manipulation because they learn the implicit density functions of the data they are trained on and create an unsupervised separation of semantic features (such as gender, age, etc) \citep{Goodfellow2014}. When the network is trained on images of faces, the latent space is transformed by the generator into a point in the manifold of realistic faces \citep{Goodfellow2014,Karras2019}. By moving the point along a vector in the manifold, we can manipulate the image along certain facial attributes while maintaining face identity \citep{Pumarola2018}, presenting a continuous mapping of these features, which would be impossible to obtain from any dataset.

In this initial proof of principle, we test whether the approach can identify an individual’s own face by manipulating the age and emotion of an original photograph and considering the ground truth to be the non-manipulated image in the space.

\section{Methods}

Our algorithm has four main components: 1) a pre-trained GAN to sample from the face manifold; 2) a face encoder that allows us to obtain the latent representation for any real face in order to find its position in the manifold; 3) learned features directions from the latent space to manipulate images; 4) a Bayesian optimization algorithm that efficiently samples the space.

\subsection{GAN for object generation}
GANs use an adversarial process to learn to generate realistic samples of the data it is trained on. After training a generator G and a discriminator D in a two-player minimax game with value function V(D, G), the generator is able to generate realistic samples from the original data distribution \citep{Goodfellow2014}.

\begin{equation} min\, max \,V(D,G) =E(x)[log(D(x))]+E[log(1-D(G(z))]  \end{equation}

Here, we used the StyleGAN \citep{Karras2019}, pre-trained on the Flickr-Faces-HQ dataset (FFHQ). Our input is set to the intermediate latent space of the network $( \,\textit{dlatent} \in \mathbb{R}^{18,512} )$ \, since it provides a more disentangled representation of the features, as it is not constrained to the probability distribution of the training data.

\subsection{Latent Space encoder} \label{latentencoder}
In order to manipulate images that the generator network was not trained on (e.g., a photo of our participants' faces taken by their webcam), we need to project the image from the manifold of face images, towards the network’s latent space. GANs consist of multiple layers of non-linear transformations, which makes it challenging to invert the network \citep{Creswell2019,Abdal2019}. Using the \textit{styleganencoder}\footnote{github.com/Puzer/stylegan-encoder}, we do so by projecting both the image we want to transform and the generated images into a common feature space of a perceptual model - \textit{conv3\_2} of the VGG16 pre-trained on the ImageNet dataset \citep{Simonyan2015}. Then, the latent values are optimized through Gradient Descent on the perceptual loss for 500 iterations, where \textit{F(I)}\, is the output of the feature space and \textit{I} is the image input. 

\begin{equation} L_{percept}(I_{1},I_{2}) =|| F(I_{1}) - F(I_{2})||^{2} \end{equation}

This process results in a latent representation that, when fed into the generative network, outputs an image that is near identical to the original one \citep{Bojanowski2018} (see Figure \ref{encoderexample}).

\subsection{Features across latent space}
It has been previously demonstrated that the organisation of the latent vector can be exploited to control different features of the image \citep{Radford2016,Pumarola2018}. In order to manipulate specific features while maintaining facial identity, we generate a set of images from the GAN and label them according to a binary categorical feature (e.g., happy vs neutral). A logistic regression is then fitted to the image’s latent representations to predict a categorical feature. Furthermore, the coefficients (\textit{c}) from the fitted regression are added to the latent representation to shift the generated image towards the fitted features, with its degree of change controlled by a scalar magnitude multiplied by the whole vector. In our research, we explore a bounded magnitude, $x \{x \in \mathbb{R} : -2 \leqslant x \leqslant 2\}$ for each feature.

\begin{equation} dlatent_{new}, =\, dlatent\, +\, c * x  \end{equation}

\subsection{Bayesian optimization}
Bayesian optimization is a powerful iterative method to efficiently obtain the extrema of target functions that are expensive to evaluate \citep{Mockus1994}. The approach employs a Bayesian statistical model to estimate the target function \textit{f(x)} according to its posterior distribution, and an acquisition function that proposes the point in the face space to be sampled in the next iteration. The statistical model used is a Gaussian process and at each iteration, has its posterior distribution and variance updated based on the new sample. The acquisition function measures the mean and confidence interval of the distribution of the current posterior distribution. Based on this, it identifies the most informative point to be sampled in the next iteration in order to find \textit{f}’s global maxima. We chose as the acquisition function the \textit{upper confidence bound} (UCB) \citep{Cox1992}. The level of exploration is controlled by the parameter $\kappa$. A high value of  $\kappa$ will privilege exploration of the space \citep{Brochu2010}, eventually resulting in active learning to map the whole space. A lower value of $\kappa$ will make the algorithm efficiently find the maxima of the space.

\begin{equation} UCB(x) = \mu(x) + \kappa\sigma(x)
  \end{equation}

Bayesian optimization is ideally suited to perform  optimal stimulus selection in the context of neuroscientific research \citep{Lorenz2016,Lorenz2017,Lorenz2018} because a) evaluating all possible stimuli is not feasible with human participants, b) the target functions are unknown (e.g., structure, concavity, number of maxima or linearity), c) the sampled values are “derivative-free”, limiting the use of any gradient descent approaches, and d) the neural or behavioural samples will inherently be affected by stochastic noise. 


Our set $A$, the face space, is a hyper-rectangle $\{x \in \mathbb{R}^{d} : a_{i} \leqslant x_{i} \leqslant b_{i}\}$, where each dimension d manipulates one disentangled facial feature across its axis. The choice of features is flexible and should be adapted depending on the specific research question. This is the space that the Bayesian optimization will navigate, where a point in the space represents an image generated by the generator network with the latents ($dlatent + \Sigma\, c_{i}*x_{i}$). 

\subsection{The framework}
Our proposed framework is a combination of these four algorithms to automatically explore the face space and find the maximum of a target function that varies across the chosen feature manipulations. It can use as input any real face, which is automatically encoded into its latent representation by minimizing the differences of the generated image and the real face in a perceptual space. The target function is first evaluated after 5 burn-in samples, uniformly chosen at random to fill the space. Each point is converted to an image through the generator network, displayed to the participant and the response is measured and fed back to the algorithm. After the initial five iterations, the loop is closed by the Bayesian optimization algorithm automatically choosing the points to sample for the next 20 iterations, with each point sample following the same steps as before (see Algorithm \ref{pseudo}). 

\begin{figure}
\includegraphics[scale=0.40]{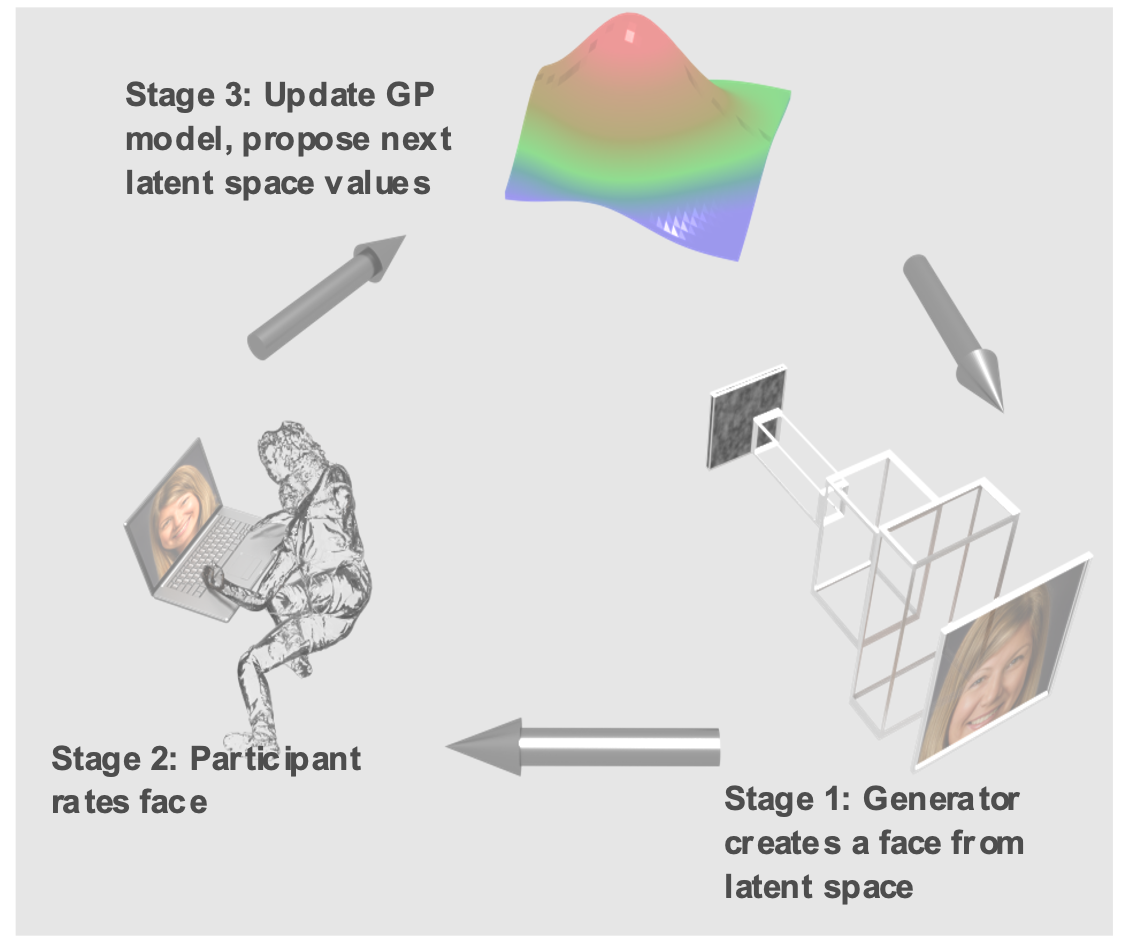}
\centering
\caption{Schematic representation of the framework.}
\end{figure}

We use a combination of a Matérn 5/2 kernel and a white noise kernel to allow for noisy inputs \citep{Rasmussen2004}. The algorithm is wrapped around a GUI that is run with Google Colab to take advantage of Google hardware to run the generator network. Its automated and flexible process allows any user with an internet connection to run the software from end-to-end.

\section{Proof of concept study}
To demonstrate the framework, we conducted a web-based, behavioural study with 30 participants (14 female, mean $\pm$ sd age: 31.33 $\pm$ 13.94 years) in which they had to rate manipulated photos of themselves.  The aim was to quickly identify the face stimuli that maximally resembled their original, non-manipulated photo.
For this, we defined a face space composed of two dimensions, age and emotion, where each axis is a linear variation of these features across the latent vector. A negative value corresponded to an older version in the age axis and to an angrier version in the emotion axis. Each participant took a photo that was encoded into the latent space. At each of the 25 iterations, participants were shown a manipulated image of their original photo and were instructed to rate the  similarity between them (0 being nothing alike and 10 being exactly like their original photo). Each manipulated image corresponded to the transformation associated with the point sampled in the space. This study design allowed us to benchmark the algorithm's performance against a known ground-truth (the non-manipulated image in the space). In addition, for six participants, we conducted further runs to assess the algorithm's test-retest reliability (first and second run) and compare the algorithm's performance against random search (third run).

The results showed that the maximum is more dispersed on the age axis than on the emotion axis, although the median response tends to be near the origin of the space (median $\pm$ sd for emotion: -0.04 $\pm$ 0.15; age: -0.06 $\pm$ 0.30). The test-retest reliability analysis showed a high intra-subject spatial correlation (mean Pearson correlation coefficient across participants  $\pm$ sd: 0.76 $\pm$ 0.14); higher than the mean inter-subject correlation between participants' response patterns (0.64 $\pm$ 0.19). For further analysis see Appendix \ref{aresults}.

\begin{figure}
\includegraphics[scale=0.4]{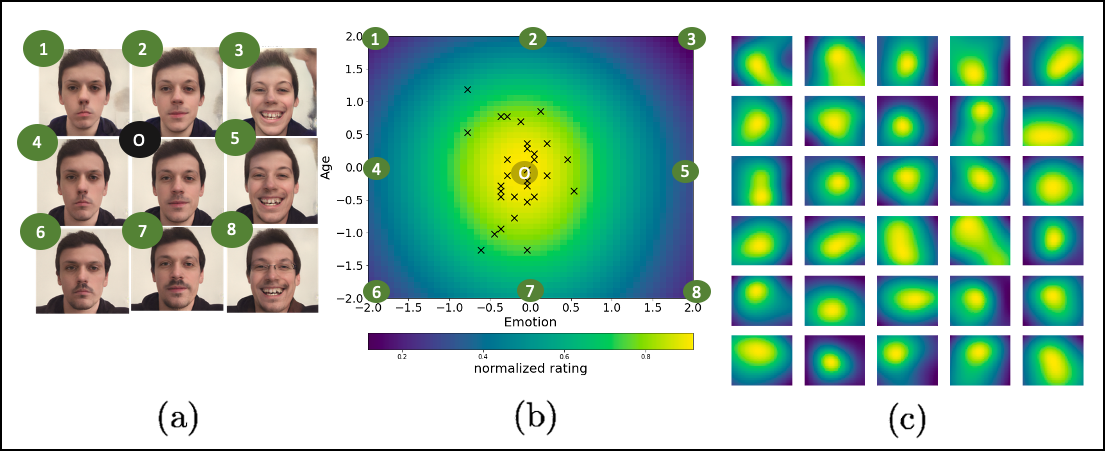}
\caption{ (a)  Images displaying the extremes of the space (1-8) and the origin point (O). (b) Mean response of the 30 participants. Each individual maximum response is marked with a cross. (c) Individual response of each participant across the space.}
\centering
\end{figure}

\section{Discussion}
This paper proposes a new tool to automatically generate and manipulate face stimuli across several semantic directions in a well-controlled manner. We showed that after only a few iterations we can identify the optimal face stimulus to maximise a target response and can accurately predict the individual's response across the entire face space. Importantly, we showed that response patterns are more stable within individuals than across participants. This  suggests that there might be indeed inter-individual response patterns. Further, high intra-subject reliability is a critical prerequisite for bringing this method out of the lab. 

This approach is relevant for a wide range of disciplines interested in an individual’s response to faces (e.g., neuroscience, psychology, psychiatry, marketing). In a clinical setting, altered response patterns to faces could be used to guide diagnosis or patient stratification for neuropsychiatric conditions known to affect face processing (e.g., Autism spectrum disorder \citep{Golarai2006}, fronto-temporal dementia). In experimental neuroscience, it allows to identify a set of face  stimuli that evoke similar brain responses but bypass effects of habituation. For psychology, it could be used to investigate how different emotions or personality traits might result in different response patterns.



The space is not limited to be 2-dimensional and there is no limitation on the types of images that can be presented. GANs have been used to learn different manifolds (e.g., houses, animal faces), which could be used to create a navigable space following the same framework. Equally, sounds could also be optimized in the same way. Regarding limitations of the stimuli, the extremes of the space will sometimes display distorted images. One reason is that we are interpolating linearly between categories in the latent space, where a non-linear transformation would be able to better capture the transition across the axis.

In conclusion, our framework offers an interesting tool for human-centred research.

\acks{This work was supported by SAPIENS Marie Curie Slowdowska Actions ITN N. 814302, the Wellcome Trust (209139/Z/17/Z), the AIMS-2-TRIALS programme funded by the Innovative Medicines Initiative (IMI) Grant No. 777394, European Union’s Horizon 2020),
the Medical Research Council (Ref: MR/R005370/1) and the Wellcome/EPSRC Centre for Medical Engineering (Ref: WT 203148/Z/16/Z) and would like to acknowledge support from the Data to Early Diagnosis and Precision Medicine Industrial Strategy Challenge Fund, UK Research and Innovation (UKRI).}

\vskip 0.2in
\bibliography{sample}


\newpage

\appendix
\section{Latent space encoding and new direction examples}
\begin{figure}
\centerline{\includegraphics[scale=0.4]{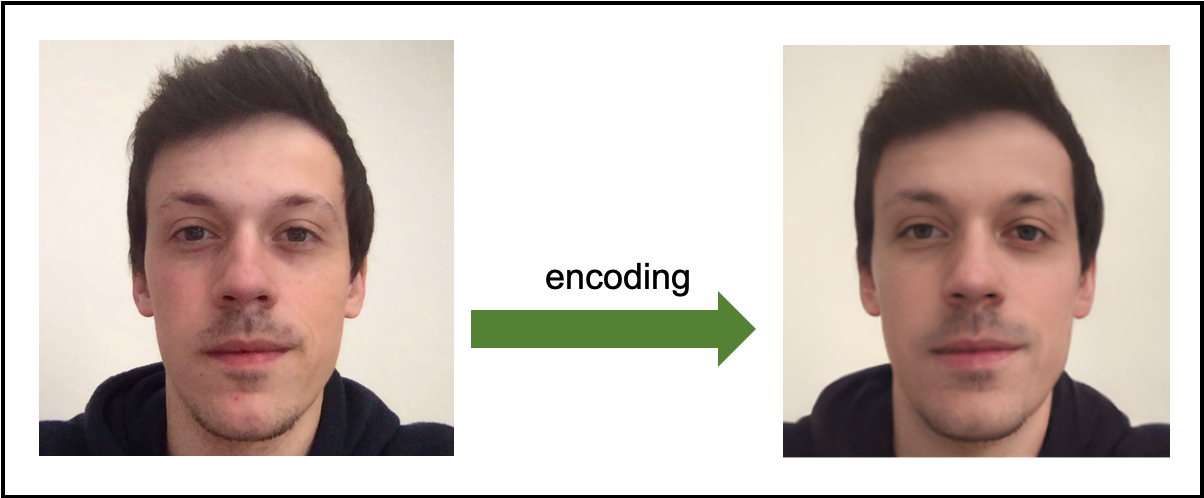}}
\caption{  Latent Space encoding example.}
\centering
\label{encoderexample}
\end{figure}

\begin{figure}
\centerline{\includegraphics[scale=0.5]{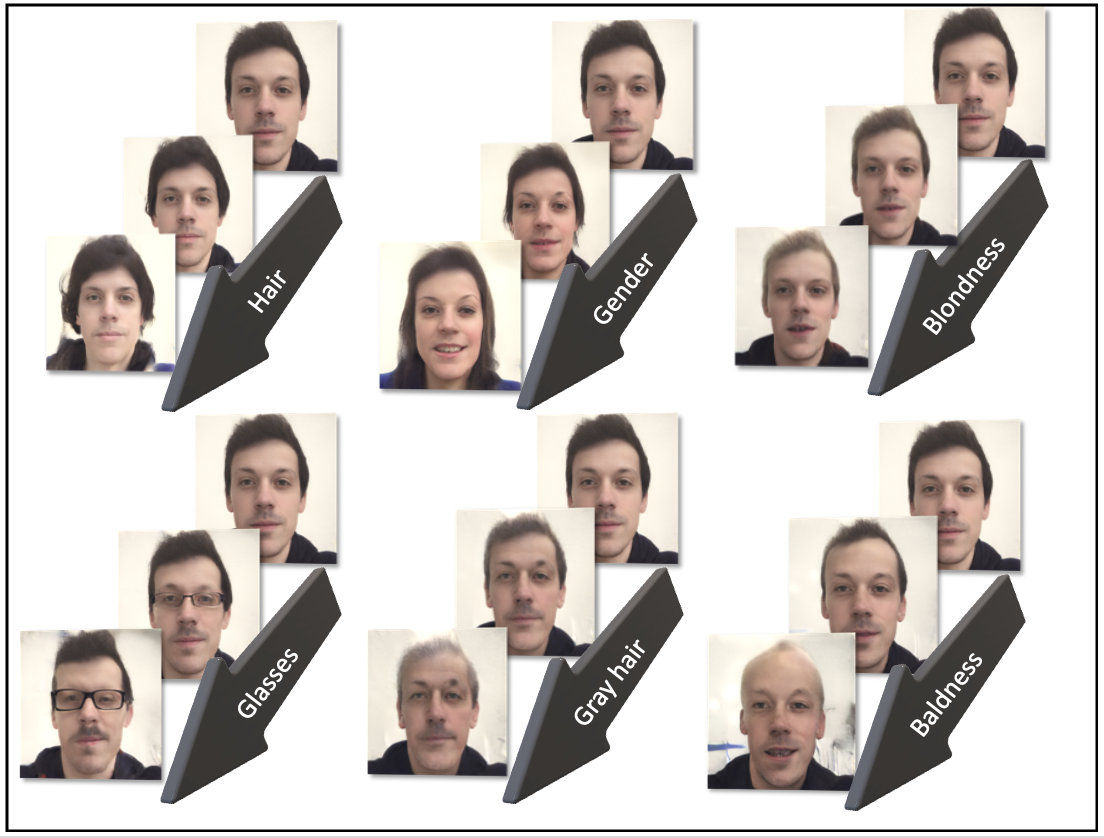}}
\caption{Examples of axis in the latent space resulting in different semantic transformations of the generated image.}
\centering
\label{directionexample}
\end{figure}

As a demonstration of the reliability of the encoder, we present the results of an encoding from the face space to the latent space in figure \ref{encoderexample}.
The  image on  the  left  is  a  photograph of the author.   The image on the right is generated from a multidimensional datapoint in the latent space that was chosen by optimizing the latent values through gradient descent on a perceptual loss (see Section \ref{latentencoder}) This results were obtained with 500 epochs, which took 7 minutes to run on Google Colab. 
By learning the latent representation of a figure, we can manipulate it by applying linear transformations in the latent space to transform semantic attributes of the original image. The constructed space is not limited to the semantic directions described in the paper as Figure \ref{directionexample} demonstrates. \\
 \begin{algorithm}
\caption{Framework pseudo-code}\label{pseudo}
\begin{algorithmic}[1]
\Procedure{Burn-in}{}

\While {$n \leqslant 5$}
    \State randomly sample $x_{n}$ from $A$.
    \State $dlatent_{new} = dlatent\, +\, c*\, x_{n}$.
    \State Run $G(dlatent_{new}) = image\_stimulus$.
    \State Observe $y_{n} = f(x_{n})$.
    \State Increment $n$.
\EndWhile

\EndProcedure

\Procedure{B.O.}{}
\While {$n \leqslant 25$}
    \State Update posterior probability distribution on $f$ using sampled points.
    \State Let $x_{n}$ be the maximizer of the acquisition function.
    \State $dlatent_{new} = dlatent\, +\, c*\, x_{n}$.
    \State Run $G(dlatent_{new}) = image\_stimulus$.
    \State Observe $y_{n} = f(x_{n})$.
    \State Increment $n$.
\EndWhile
\State \Return sampled point with largest posterior mean

\EndProcedure

\end{algorithmic}
\end{algorithm}

\section{Further Analysis of the data}
\label{aresults}
\begin{figure*}[!b]
\centerline{\includegraphics[scale=0.4]{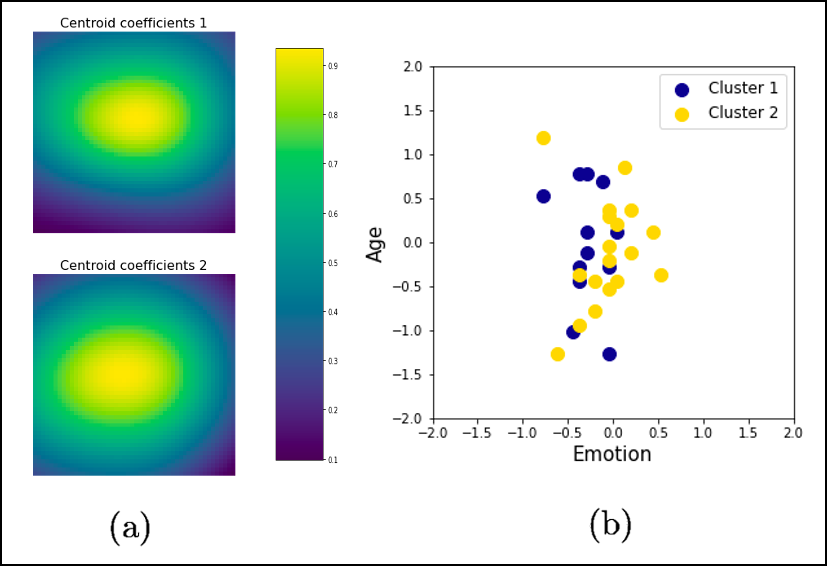}}
\caption{ Cluster Analysis using K-means clustering on the full space of the participants’ evaluations. (a) Centroids for the two clusters identified. The first centroid captures a higher dispersion towards positive values of age (younger images) and the second centroid captures a higher dispersion towards negative values of emotion (angrier images). (b) Maximum of each participant labelled according to its cluster.}
\centering
\label{cluster}
\end{figure*}

\begin{figure}
\centerline{\includegraphics[scale=0.4]{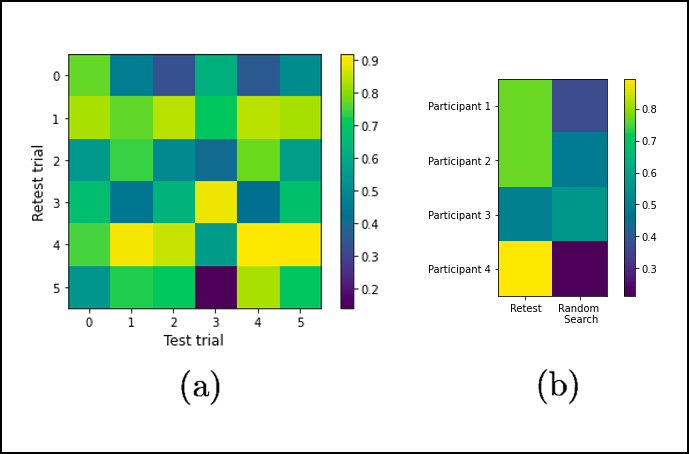}}
\caption{ (a) Similarity matrix between the target space of different runs of the same participant (matrix diagonal) and between different participants. The mean Pearson correlation between trials of the same participant is 0.76, where the correlation between trials of different participants is 0.64. (b) Correlation matrix between the predicted target space of the first run and the second run (first column) and between the first run and a run using random search across the space instead of Bayesian optimization (second column).}
\centering
\label{correlation}
\end{figure}

\begin{figure}
\centerline{\includegraphics[scale=1]{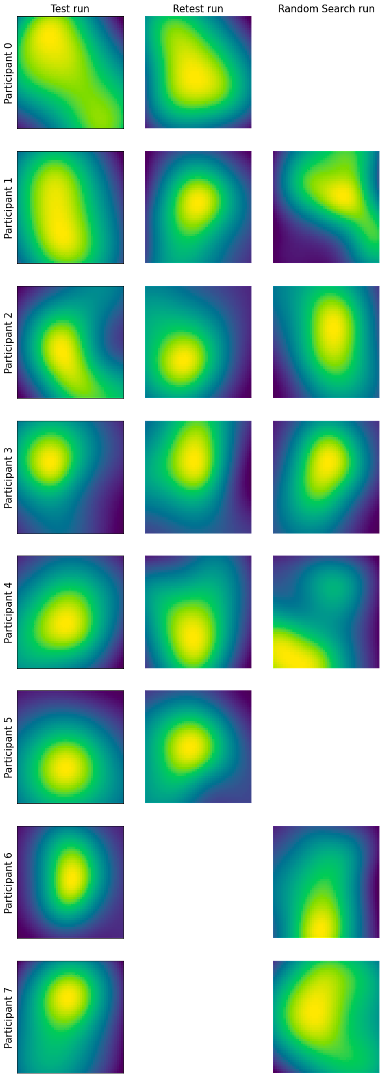}}
\caption{ Individual response of all participants that performed the algorithm twice (6 participants) and that performed the evaluation with random sampling across the space (6) modelled using Gaussian processes with a fixed kernel.}
\centering
\label{response}
\end{figure}

Test-retest analysis indicated that intra-individual participant patterns in different runs correlated more than with the patterns of other individuals. This result seems to sustain the argument that the framework might be able to capture personalized responses on self-perception. To analyse this further, k-means clustering was performed for two clusters on the full space predictions of the participant's response. The silhouette score was of 0.17. The results are displayed in Figure \ref{cluster}. 
An analysis of the correlation of the test runs with the re-test runs and a run not using the sampling algorithm (i.e. using a random search algorithm) shows that the correlation between the two former (mean $\pm$ sd.: 0.74 $\pm$ 0.14) is higher between the test and the random-search patterns (mean $\pm$ sd.: 0.41 $\pm$ 0.13). The results are presented in Figures \ref{correlation} and \ref{response}.

\end{document}